\documentclass{article}

\usepackage[preprint]{neurips_2020}

\usepackage{amsmath,amsfonts,bm}

\def\eqref#1{equation~\ref{#1}}

\def\1{\bm{1}}

\DeclareMathAlphabet{\mathsfit}{\encodingdefault}{\sfdefault}{m}{sl}
\SetMathAlphabet{\mathsfit}{bold}{\encodingdefault}{\sfdefault}{bx}{n}

\usepackage{hyperref}
\usepackage{url}
\usepackage{amsfonts}
\usepackage{amssymb}
\usepackage{tasks}
\usepackage{graphicx}
\usepackage{xcolor}
\usepackage[ruled,linesnumbered,vlined]{algorithm2e}
\usepackage{xspace}
\newcommand\rfs{{\em reasoning-for-sure}\xspace}

\newcommand\syllogism[3][]{%
  \begin{center}
  \def\tmp{#1}%
  \ifx\tmp\empty\else(#1)\quad\fi
  \begin{tabular}{@{}l@{}}#2\\\hline#3\quad$\therefore$
  \end{tabular}
  \end{center}
}
\title{An AI Monkey Gets Grapes for Sure -- Sphere Neural Networks for Reliable Decision-Making}

\author{%
Tiansi Dong\\
Computer Laboratory\\
       University of Cambridge\\
       15 JJ Thomson Ave, Cambridge, UK\\
\texttt{td540@cam.ac.uk} 
 \And
Henry He\\
Phillips Academy Andover\\
180 Main Street, Andover\\ Massachusetts, USA\\
\texttt{hhe26@andover.edu} 
 \And
Pietro Liò\\
Computer Laboratory\\
       University of Cambridge\\
       15 JJ Thomson Ave, Cambridge, UK\\
\texttt{pl219@cam.ac.uk} \\
 \And
Mateja Jamnik\\
Computer Laboratory\\
       University of Cambridge\\
       15 JJ Thomson Ave, Cambridge, UK\\
\texttt{mj201@cam.ac.uk} \\ 
}

\begin{document}

\maketitle

\begin{abstract}
This paper compares three methodological categories of neural reasoning: LLM reasoning, supervised learning-based reasoning, and explicit model-based reasoning. LLMs remain unreliable and struggle with simple decision-making that animals can master without extensive corpora training. Through disjunctive syllogistic reasoning testing, we show that reasoning via supervised learning is less appealing than reasoning via explicit model construction. Concretely, we show that an Euler Net trained to achieve 100.00\% in classic syllogistic reasoning can be trained to reach 100\% accuracy in disjunctive syllogistic reasoning. However, the retrained Euler Net suffers severely from catastrophic forgetting (its performance drops to 6.25\% on already-learned classic syllogistic reasoning), and its reasoning competence is limited to the pattern level. We propose a new version of Sphere Neural Networks that embeds concepts as circles on the surface of an $n$-dimensional sphere. These Sphere Neural Networks enable the representation of the negation operator via complement circles and achieve reliable decision-making by filtering out illogical statements that form unsatisfiable circular configurations. We demonstrate that the Sphere Neural Network can master 16 syllogistic reasoning tasks, including rigorous disjunctive syllogistic reasoning, while preserving the rigour of classical syllogistic reasoning. We conclude that neural reasoning with explicit model construction is the most reliable among the three methodological categories of neural reasoning.  
\end{abstract}

\section{Introduction}
\label{sintro}

Reliable decision-making is crucial in high-stakes applications. Although LLMs have achieved unprecedented success in many ways, exemplified in human-like communication~\citep{chatgpt_nature2023}, playing Go~\citep{alphaGo2017,alphaGo2020}, predicting complex protein structures \citep{AlphaFold3}, or weather forecasting~\citep{DeepMindWeather24}, they still make errors in simple reasoning~\citep{Mitchell_science.adj5957}. In addition, LLMs are prone to making accurate predictions with incorrect explanations \citep{creswell2022selectioninference,zelikman2022star,AIDeception2024}, and have not yet achieved the reliability necessary for high-stakes applications, i.e., decision-making in biomedicine~\citep{syllobio2025}.

In some cases, syllogistic reasoning may appear deceptively simple, such as in the following: \syllogism{All members of the Diseases of Hemostasis pathway are members of the general Disease pathway.\\ Gene GP1BB is a member of Diseases of hemostasis pathway.}{Gene GP1BB is a member of Disease pathway.}  However,   \cite{Eisape2024} show that although LLMs perform better than average humans in syllogistic reasoning, their accuracy remains limited to around 75\%, and larger models do not consistently outperform smaller ones. 
\cite{syllogism24} come to the convergent conclusions that, in abstract reasoning such as syllogism, LLMs may achieve above-chance performances in familiar situations but exhibit numerous imperfections in less familiar ones. \cite{syllobio2025} tested LLMs in several types of generalised syllogistic reasoning, e.g., {\em generalised modus ponens}, {\em disjunctive syllogism}, in the context of high-stakes biomedicine, and found that zero-shot
LLMs achieved an average accuracy between 70\% on {\em generalised modus ponens} and
23\% on {\em disjunctive syllogism}. Crucially, both zero-shot and few-shot LLMs demonstrated pronounced sensitivity to surface-level lexical variations. 

Despite these limitations, evaluating the reasoning performance of LLMs on syllogistic tasks (and beyond) can provide insights into the origins of (human) rationality~\citep{syllogism24} and help identify alternative methods for reliable neural reasoning and decision-making. LLMs acquire human-like communication and reasoning abilities by training on large-scale linguistic corpora. However, proficiency in communication does not necessarily equate to proficiency in reasoning~\citep{language4commu}. On the contrary, reasoning and decision-making are abilities that do not necessarily depend on extensive language acquisition. 
For example, clever monkeys can get grapes through disjunctive syllogistic decision-making~\citep{Ferrigno21}, suggesting that syllogistic reasoning can be elicited through visual–spatial inputs, independent of linguistic abilities. 
\begin{figure} 
\centering
\includegraphics[width=1\textwidth]{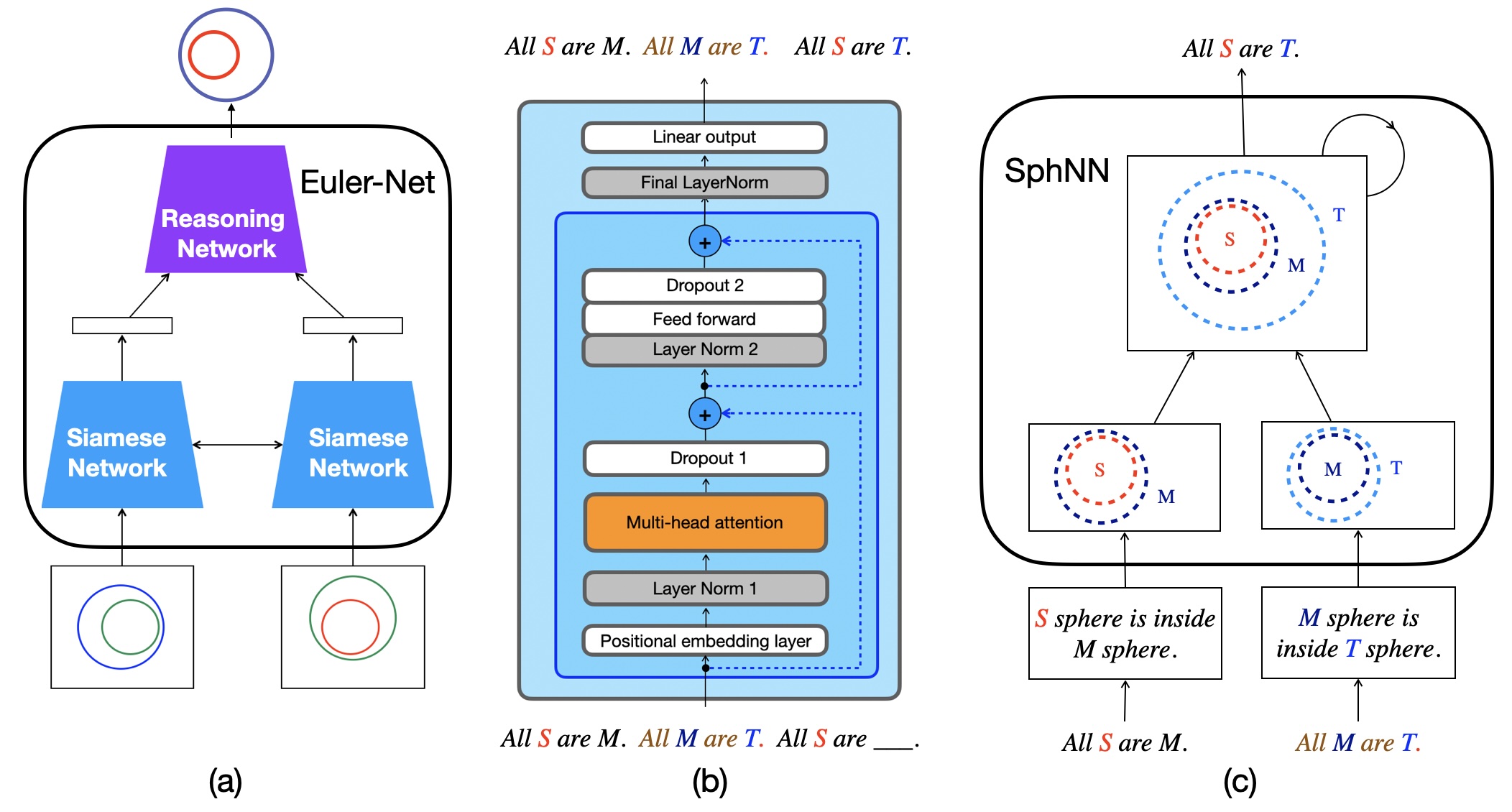}
\caption{(a) The architecture of Euler Net. The inputs are two images, representing two premises of Aristotelian syllogistic reasoning. The green circle inside the blue circle represents ``all green are blue''; the red circle inside the green circle represents ``all red are green''. Its output represents ``all red are blue''. (b) The transformer component of LLMs learns to predict the missing word. (c) The architecture of Sphere Neural Networks (SphNN) for syllogistic reasoning. SphNN transforms syllogistic statements into spatial statements between spheres, constructs a unified configuration, and draws conclusions by inspecting the constructed configuration.}
\label{3nets}
\end{figure}
While recent evaluations indicate that LLMs remain unable to achieve robust syllogistic reasoning, in this work, we revisit \citep{Ferrigno21}'s experiments to seek alternative neural architectures and neural reasoning methods. We distinguish three categories of neural networks for syllogistic reasoning (illustrated in Figure~\ref{3nets}) as follows:
\begin{enumerate}\item {\em Supervised networks with image inputs}. Inspired by the structural similarity between deep Convolution Neural Networks (CNN)~\citep{resnet} and visual cortex~\citep{Yamins16}, Euler Nets are special CNNs that perform Aristotelian syllogistic reasoning with Euler diagrammatic-styled image-inputs~\citep{WangJL18,WangJL20}. 
\item {\em Supervised networks with linguistic inputs}.
The popular neural network architecture is the Transformer~\citep{Vaswani17}, which underpins LLMs~\citep{palm22023,llama22023,openAI23,mistral7b,mistral23}. The basic training method is masked word prediction~\citep{build-llms24} using Transformers. Given the text {\em all Greeks are human. all humans are mortal. therefore, all Greeks are \_\_\_ }, LLMs are trained to predict {\em mortal}. 

\item  {\em Neural networks that reason through explicit model construction}. Sufficient empirical experiments advocate the model theory for reasoning that reasoning is a process of constructing and inspecting mental models~\citep{LairdByrne91,knauf03,GoodwinLaird05,Knauff09,Knauff13}. 
In line with the mental model theory, Sphere Neural Networks perform Aristotelian syllogistic reasoning by constructing Euler diagrams in the Euclidean or Hyperbolic space~\citep{djl2024sphere,djl2025}. They are proven to achieve the rigour of symbolic-level logical reasoning.
\end{enumerate}

In this paper, we compare reasoning performances between Category 1 {\em supervised neural networks with image inputs} and Category 3 {\em neural networks through explicit model construction} (in the Appendix, we report the performance of OpenAI GPT-5 and GPT-5-nano in syllogistic reasoning). We challenge them on various kinds of syllogistic-style decision-making, which are the core of many high-stakes applications, e.g., legal judgments~\citep{LJA23,law23}, medical diagnoses~\citep{dm10,pl2023}. 


The contributions of this paper are multifold: (1) by defining Euler diagrams as circles on the surface of an $n$-dimensional sphere, we
enhance the representation power of Sphere Neural Networks~\citep{djl2024sphere,djl2025} (a representative network in Category 3) for performing 16 first-order syllogistic-styled reasoning types~\citep{critical_thinking21}, including disjunctive syllogistic reasoning; (2)  
We evaluated our Sphere Neural Network on constructing Euler diagrams with dimensions of $2, 3, 15, 30, 100, 200, 1000, 2000, 3000,$ and $10000$, and it achieved $100\%$ accuracy across all these types, showing that our Sphere Neural Network preserved it's effectiveness across all types of classic syllogism. (3) We repurposed Euler Net~\citep{WangJL18,WangJL20} (a representative network in Category 1) to perform disjunctive syllogistic reasoning, and achieved 100\% accuracy; (4) We created new testing data to challenge the robustness of Euler Net and demonstrated that its reasoning performance is restricted by input patterns. Our experiment results favour Category 3 networks for reliable reasoning and decision-making.  
 


The rest of the paper is structured as follows: In Section~\ref{mk}, we revisit the design of \cite{Ferrigno21}'s monkey experiments as background for our experiments; In Section~\ref{related_work}, we survey syllogistic-styled reasoning and some related work; In Section~\ref{xSphNN}, we present our version of Sphere Neural Networks for various syllogistic reasoning; In Section~\ref{exp}, we conduct a comparative study between Sphere Neural Network and Euler Net and report our experiment results. In Section~\ref{con}, we conclude our work and list potential future research directions.
\begin{figure} 
\centering
\includegraphics[width=1\textwidth]{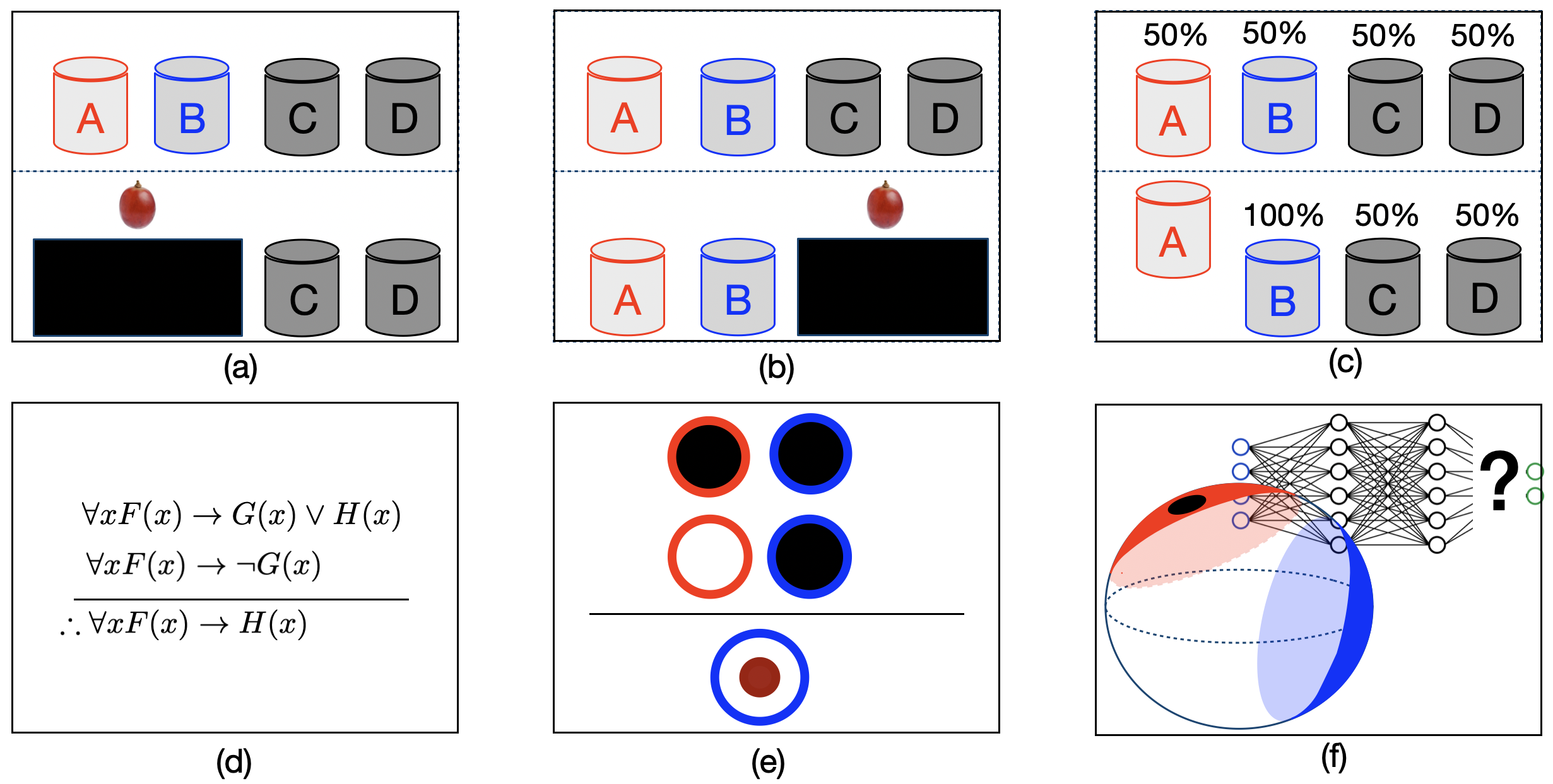}
\caption{There are four jars, A, B, C, and D, in front of monkeys. (a) A blackboard covers jars A and B. A grape is dropped into one of jars A or B behind the blackboard; (b) Another blackboard covers jars C and D. Another grape is dropped into one of jars C or D behind the blackboard; (c) After removing the blackboards, monkeys will see four jars, each having $50\%$ probability to contain a grape. If jar A is empty, clever monkeys will lift jar B. (d) The decision process of the clever monkeys can be abstracted as disjunctive syllogistic reasoning; (e) a diagrammatic representation for disjunctive syllogistic reasoning; (f) sphere neural networks and supervised neural networks may simulate disjunctive syllogistic reasoning.}
\label{intro}
\end{figure}
\section{Clever Monkeys can do disjunctive syllogistic reasoning} 
\label{mk}

\cite{Ferrigno21} conducted a series of experiments with monkeys, showing that clever monkeys can perform disjunctive syllogistic reasoning. The process can be briefly described as follows: Researchers put four jars in front of monkeys and hide the first two jars behind a board. Then, they drop a grape into one of the two jars. Thus, the monkeys know one of the two jars has the grape, but do not know which one. Then, they repeat the process for the last two jars. Thus, from the monkeys' perspective, each jar has a $50\%$ chance of having a grape. Subsequently, researchers reveal the contents of the first jar; a grape is given to the monkey if one is present; if not, the monkey is permitted to lift one of the remaining three jars to obtain a grape. In this scenario, clever monkeys will lift the second jar, which has a $100\%$ probability of containing a grape, as illustrated in Figure~\ref{intro}. We may abstract the decision-making process as a form of disjunctive syllogistic reasoning as follows. 
\begin{tasks}(2)
\task[] 
\syllogism{$\forall x\ \text{Grape}(x) \rightarrow \text{InJarA}(x)\lor \text{InJarB}(x)$.\\ $\forall x\  \text{Grape}(x) \rightarrow \neg \text{InJarA}(x)$.}{$\forall x\  \text{Grape}(x) \rightarrow \text{InJarB}(x)$.}
\task[\\ gernalised\\ version:] 
\syllogism{$\forall x G(x) \rightarrow A(x)\lor B(x)$.\\ $\forall x  G(x) \rightarrow \neg A(x)$.}{$\forall x G(x) \rightarrow B(x)$.}
\end{tasks}
\begin{figure}[h]
\centering  
  \centering 
\includegraphics[width=1\textwidth]{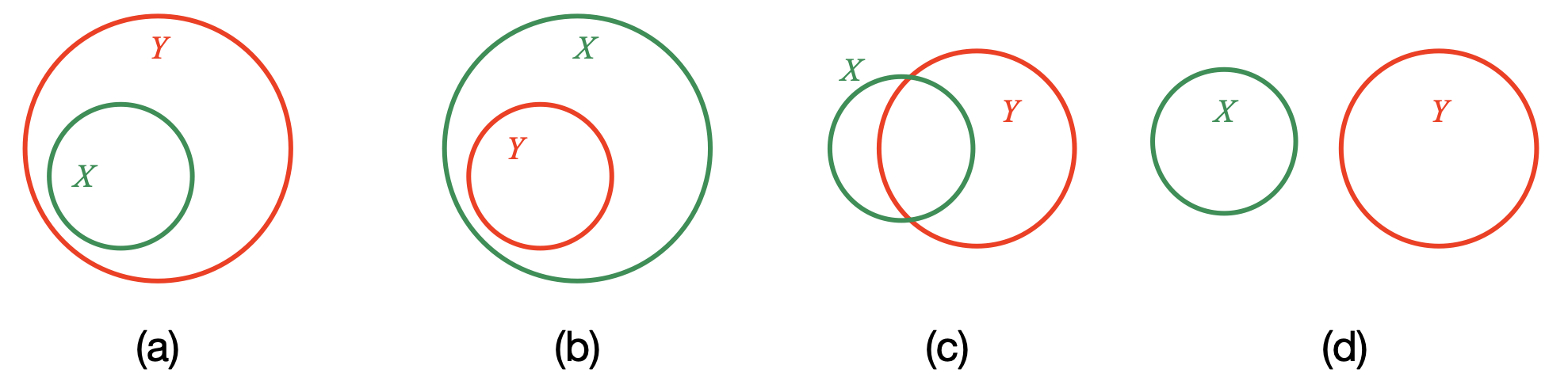}  \caption{Basic diagrammatic representation for syllogistic statements. (1) that {\em all X are Y} is represented by (a) $X\subset Y$; (2) that {\em some X are Y} is represented by (a) $X\subset Y$ or (b) $Y\subset X$ or (c) $X\cap Y\neq\emptyset$; (3) that {\em no X are Y} is represented by (d) $X\cap Y=\emptyset$; (4) that {\em some X are not Y} is represented by (b) $Y\subset X$ or (c) $X\cap Y\neq\emptyset$ or (d) $X\cap Y=\emptyset$. 
}
\label{euler_diagram}
\end{figure}
\section{Syllogistic-style reasoning}
\label{related_work}

Syllogistic reasoning \citep{jeffrey81} is a form of deductive reasoning with only two premises and three terms. A very well-known example is {\em All Greeks are humans. All humans are mortal.} $\therefore$ {\em All
Greeks are mortal.} The common concept {\em humans} in the premises establishes the relation between {\em Greeks} and {\em mortal}. Relations (or ``moods'' as used in the psychological literature) are limited to (1) {\em universal affirmative}: all $X$ are $Y$; (2) {\em particular affirmative}: some $X$ are $Y$; (3) {\em universal negative}: no $X$ are $Y$; (4) {\em particular negative}: some $X$ are not $Y$. By allowing terms to exchange places in the premises, we distinguish 256 different types of syllogistic reasoning~\citep{laird2012}. The four relations can be reduced to four basic set relations in the forms of Euler diagrams~\citep{hammer98}: (a) $X$ is part of $Y$ ($X\subset Y$), (b) $X$ contains $Y$ ($Y\subset X$), (c) $X$ partially overlaps with $Y$ ($X\cap Y\neq \emptyset$), and (d) $X$ is disjoint from $Y$ ($X\cap Y=\emptyset$), as shown in Figure~\ref{euler_diagram}.  
Euler diagrams can be constructed as a sphere configuration either in Euclidean space~\citep{djl2024sphere} or in Hyperbolic space~\citep{djl2025} and achieve the rigour of syllogistic reasoning. The method maps Set X to Sphere $\mathcal{O}_X$ and translates four syllogistic relations into four spatial relations as follows.  
\begin{itemize}
    \item ``all $X$ are $Y$'' $\Leftrightarrow$ ``sphere $\mathcal{O}_X$ is part of sphere $\mathcal{O}_Y$'', $\mathbf{P}(\mathcal{O}_X, \mathcal{O}_Y)$;
    \item ``some $X$ are $Y$'' $\Leftrightarrow$ ``sphere $\mathcal{O}_X$ does not disconnect from sphere $\mathcal{O}_Y$'', $\neg\mathbf{D}(\mathcal{O}_X, \mathcal{O}_Y)$;
    \item ``no $X$ are $Y$'' $\Leftrightarrow$ ``sphere $\mathcal{O}_X$ disconnects from sphere $\mathcal{O}_Y$'', $\mathbf{D}(\mathcal{O}_X, \mathcal{O}_Y)$;
    \item ``some $X$ are not $Y$'' $\Leftrightarrow$ ``sphere $\mathcal{O}_X$ is not part of sphere $\mathcal{O}_Y$'', $\neg\mathbf{P}(\mathcal{O}_X, \mathcal{O}_Y)$.
\end{itemize} 

\begin{figure}
\centering
\includegraphics[width=1\textwidth]{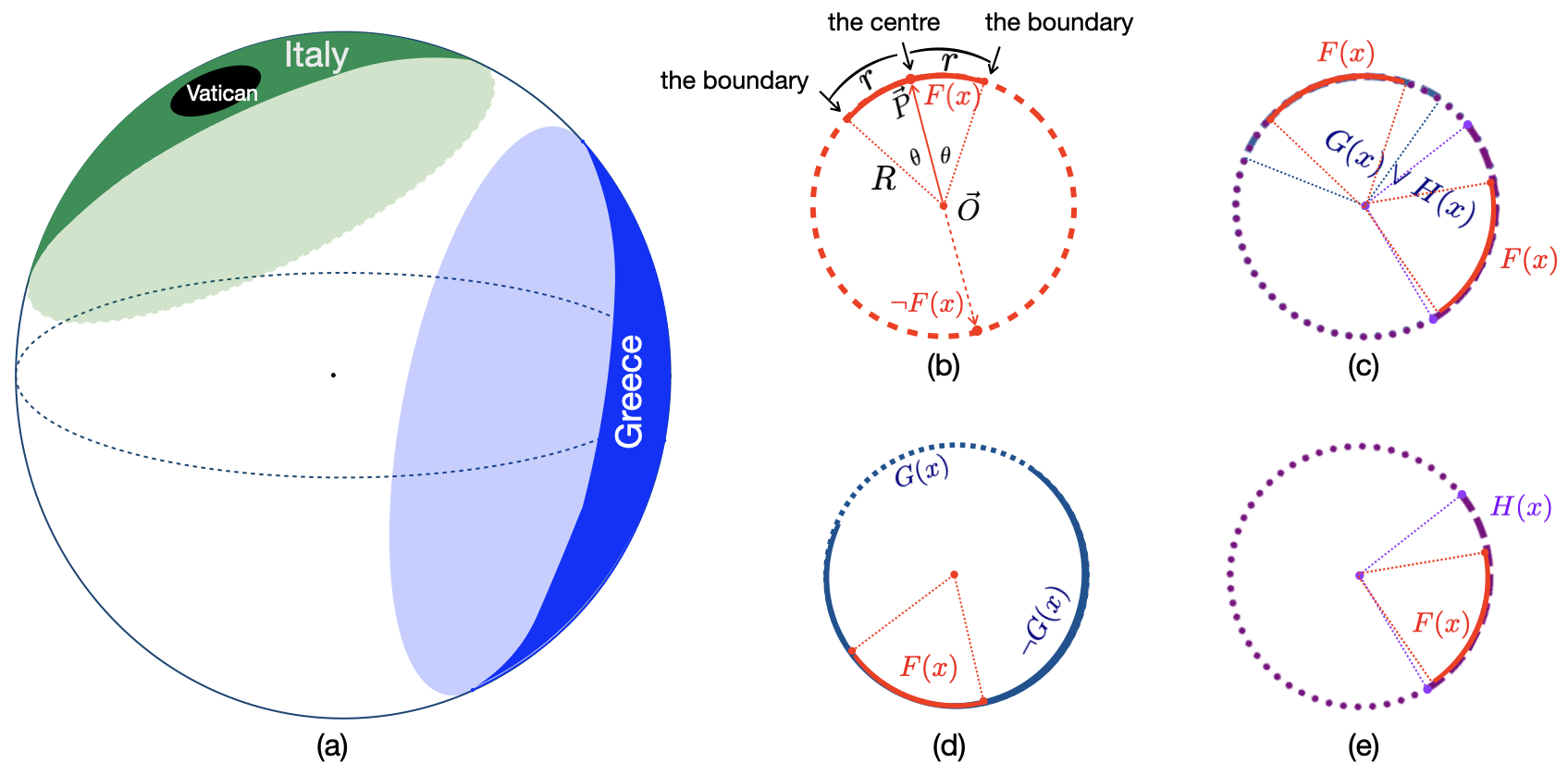}
\caption{(a) An Euler diagram on the surface of a 3-d sphere. 
(b) The surface of a 2-d sphere is a circle. A sphere on this surface is a curve, e.g., $F(x)$, $\neg F(x)$; (c) $\forall x\cdot F(x)\rightarrow G(x)\lor H(x)$: the $F(x)$ arc is part of $G(x)$ arc or $H(x)$ arc; (d) $\forall x\cdot F(x)\rightarrow \neg G(x)$: $F(x)$ arc is part of $\neg G(x)$ arc; (e) $\forall x\cdot F(x)\rightarrow H(x)$: $F(x)$ arc is part of $H(x)$ arc.}
\label{fig:arc_sphere}
\end{figure}
 
Embedding data on a spherical surface efficiently can solve rotation-invariant learning problems~\citep{sphcnn18,sphcnn23a}. Here, we embed concepts as circles on the surface of an $n$-dimensional sphere and construct configurations of circles as Euler diagrams for syllogistic reasoning. If $n=3$, these circles are cones. As an example, {\em Vatican is inside Italy. Italy disconnects from Greece. $\therefore$ Vatican disconnects from Greece.}, is illustrated visually in Figure~\ref{fig:arc_sphere}(a). 

When we draw a circle on the Earth’s surface, the line from its centre to the circumference follows a curve along a great circle of the Earth. 
Formally, let $\mathcal{O}$ be an $n$-dimensional sphere with the centre $\vec{O}$ and radius $R$, $\vec{P}$ be a point on the surface of $\mathcal{O}$, $\|PO\| = R$. A circle with $\vec{P}$ as the centre and $r$ as the radius, $\bigcirc(\vec{P}, r)$, on the surface of $\mathcal{O}$, is defined as the set of points $\vec{Q}$ on the surface of $\mathcal{O}$, whose surface distance to $\vec{P}$ is less than $r$, $\arccos(\frac{\vec{P}}{\|\vec{P}\|}\cdot\frac{\vec{Q}}{\|\vec{Q}\|})R < r$. The complement of $\bigcirc(\vec{P}, r)$ is also a circle, written as $\overline{\bigcirc}(\vec{P}', r')$, where $\vec{P}' = 2\vec{O} -\vec{P}$ and $r'= (\pi - \arccos(\frac{\vec{P}}{\|\vec{P}\|}\cdot\frac{\vec{Q}}{\|\vec{Q}\|}))R$. 
When $n=2$, circles are arcs, as shown in Figure~\ref{fig:arc_sphere}(b). This definition of circles is compatible with the sphere definition in~\citep{djl2024sphere,djl2025} and can be used to create Euler diagrams for general syllogistic reasoning as follows: Let $a$ be a constant, $x$ be a variable, $F, G, H$ be predicates. 
\begin{itemize} 
    \item $F(a)$ is translated into ``Atomic Circle $a$ is part of Circle F'', Atomic Circle $a$ has the minimal radius $\epsilon>0$. Formally, $\mathbf{P}(\bigcirc_a,\bigcirc_F)$, where $r_a = \epsilon$.
    \item $\forall x F(x) \rightarrow G(x)$ is translated into ``for any Atomic Circle $x$, if $x$ is part of Circle F, $x$ is part of Circle G''. Formally, we write $\mathbf{P}(\bigcirc_F,\bigcirc_G)$.
    \item $\forall x F(x) \rightarrow G(x)\lor H(x)$. Formally, we write $\mathbf{P}(\bigcirc_F,\bigcirc_G)\lor \mathbf{P}(\bigcirc_F,\bigcirc_H)$. 
    \item $\forall x F(x) \rightarrow \neg G(x)$. Formally, we write $\mathbf{P}(\bigcirc_F,\overline{\bigcirc_G})$.
    \item $\exists x F(x) \rightarrow G(x)$ is translated into ``there is Atomic Circle $a$, if $a$ is part of Circle F, $a$ is part of Circle G''. Formally, we write $\mathbf{P}(\bigcirc_a,\bigcirc_F)\land \mathbf{P}(\bigcirc_a,\bigcirc_G)$, where $r_a = \epsilon$.
\end{itemize} 
We create circle configurations on an $n$-dimensional sphere for the following 16 types of syllogistic-styled reasoning in \citep{critical_thinking21} (Details are described in the supplementary material).
\begin{tasks}(4)
\task[1]
\syllogism{$\forall x F(x) \rightarrow G(x)$.\\ $F(a)$.}{$G(a)$.}
\task[2] \syllogism{$\forall x F(x) \rightarrow \neg G(x)$.\\ $F(a)$.}{$\neg G(a)$.} 
\task[3] \syllogism{$\forall x F(x) \rightarrow G(x)$.\\ $\neg G(a)$.}{$\neg F(a)$.}
\task[4] \syllogism{$\forall x F(x) \rightarrow \neg G(x)$.\\ $G(a)$.}{$\neg F(a)$.}
\end{tasks}
\begin{tasks}(3)
\task[5] \syllogism{$\forall x F(x) \rightarrow \neg G(x)$.}{$\forall x G(x) \rightarrow \neg F(x)$.}
\task[6] \syllogism{$\forall x F(x) \rightarrow G(x)$.}{$\forall x \neg  G(x) \rightarrow \neg F(x)$.}
\task[7] \syllogism{$\forall x F(x) \rightarrow G(x)$.\\ $\forall x G(x) \rightarrow H(x)$.}{$\forall x F(x) \rightarrow H(x)$.}
\task[8] \syllogism{$\forall x F(x) \rightarrow \neg G(x)$.\\ $\forall x \neg G(x) \rightarrow H(x)$.}{$\forall x F(x) \rightarrow H(x)$.}
\task[9] \syllogism{$\forall x F(x) \rightarrow G(x)$.\\ $\forall x \neg H(x) \rightarrow \neg G(x)$.}{$\forall x F(x) \rightarrow H(x)$.}
\task[10] \syllogism{$\forall x F(x) \rightarrow \neg G(x)$.\\ $\forall x \neg H(x) \rightarrow  G(x)$.}{$\forall x F(x) \rightarrow H(x)$.}
\task[11] \syllogism{$\forall x F(x) \rightarrow G(x)$.\\ $\exists x  H(x) \rightarrow \neg G(x)$.}{$\exists x H(x) \rightarrow \neg F(x)$.}
\task[12] \syllogism{$\forall x \neg F(x) \rightarrow G(x)$.\\ $\exists x  H(x) \rightarrow \neg G(x)$.}{$\exists x H(x) \rightarrow F(x)$.}
\task[13] \syllogism{$\forall x F(x) \rightarrow G(x)\lor H(x)$.\\ $\forall x  F(x) \rightarrow \neg G(x)$.}{$\forall x F(x) \rightarrow H(x)$.}
\task[14] \syllogism{$\forall x F(x) \rightarrow G(x)\lor H(x)$.\\ $\forall x  G(x) \rightarrow \neg F(x)$.}{$\forall x F(x) \rightarrow H(x)$.}
\task[15] \syllogism{$\forall x F(x) \rightarrow G(x)\lor H(x)$.\\ 
$\forall x  G(x) \rightarrow J(x)$.\\
$\forall x  H(x) \rightarrow J(x)$.}{$\forall x F(x) \rightarrow J(x)$.}
\task[16] \syllogism{$\forall x F(x) \rightarrow G(x)\lor H(x)$.\\ 
$\forall x  J(x) \rightarrow \neg G(x)$.\\
$\forall x J(x) \rightarrow \neg H(x)$.}
{$\forall x F(x) \rightarrow \neg J(x)$.}
\end{tasks} 

\begin{figure} 
  \centering  
\includegraphics[width=1\textwidth]{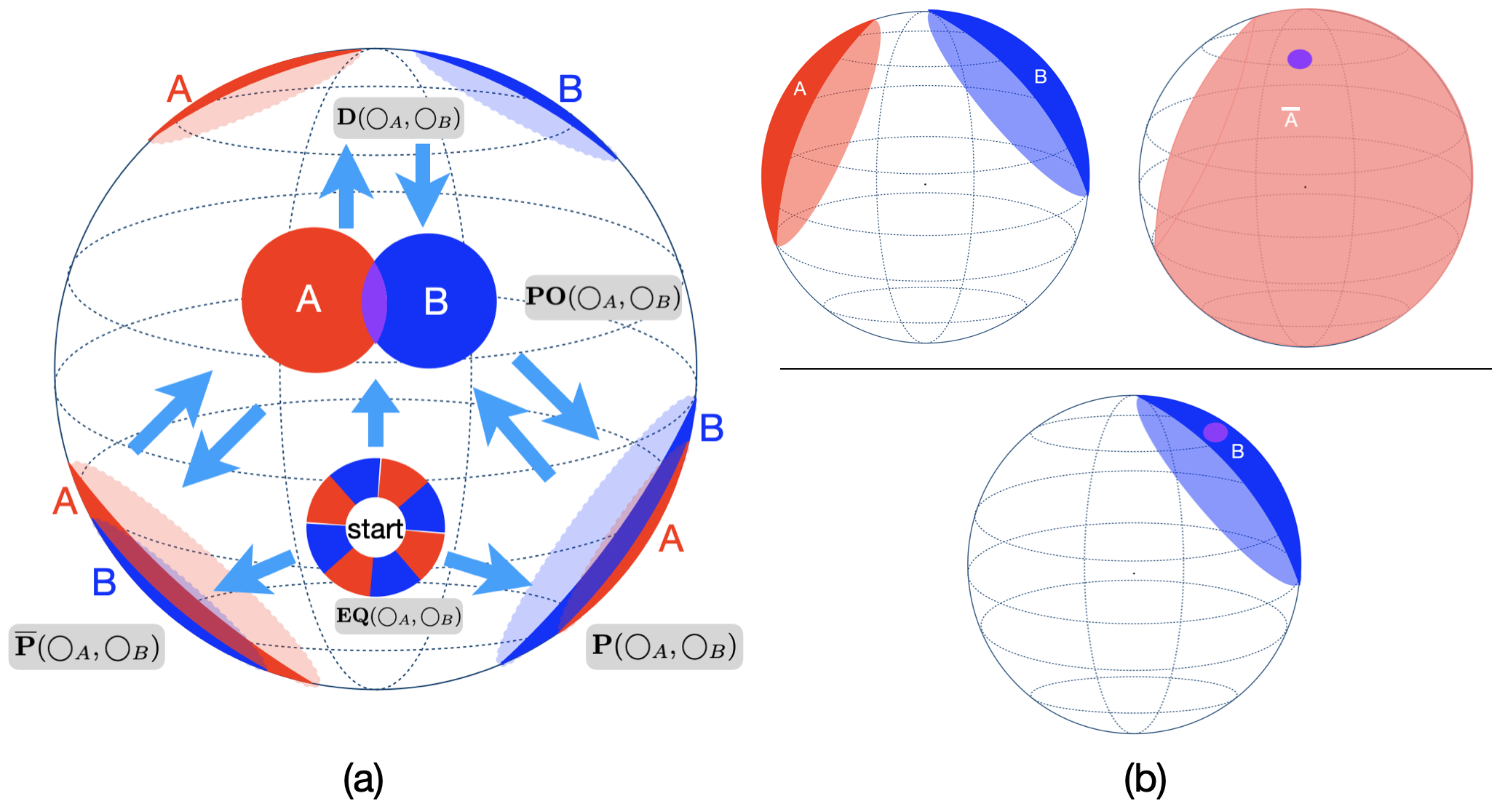}
  \caption{(a) The neighbourhood transition map between circles on the surface of a sphere. $\mathbf{PO}$ stands for ``partial overlap'', $\mathbf{EQ}$ stands for ``equal with'', $\overline{\mathbf{P}}$ stands for ``inverse part of'' or ``contain''; (b) the circle configuration of disjunctive syllogistic reasoning: (Premise 1) a grape $\bigcirc_g$ is either in jar A or in jar B, $\mathbf{P}(\bigcirc_g,\bigcirc_A)\lor\mathbf{P}(\bigcirc_g,\bigcirc_B)$; (Premise 2) the grape is not in jar A, $\mathbf{P}(\bigcirc_g, \overline{\bigcirc_A})$; (Conclusion) the grape is in jar B, $\mathbf{P}(\bigcirc_g, \bigcirc_B)$. 
  } 
  \label{monkey1}
\end{figure} 
\section{Sphere Neural Network for Reliable Disjunctive Syllogistic Reasoning}
\label{xSphNN}

Our new Sphere Neural Network (SphNN) is an extension of SphNN~\citep{djl2024sphere,djl2025} in three aspects: (1) we define spheres as circles on an $n$-d sphere; (2) we allow complement sets and define a complement circle for the complement set; (3) our SphNN determines the satisfiability of logical formula in a disjunctive normal form $f_1\lor...\lor f_n$ by explicitly constructing circle configurations for syllogistic-styled statements $f_i$ one by one, where $f_i$ is a conjunctive form $g_1\land ...\land g_m$, $g_i$ is limited to one of the forms: $\mathbf{P}(\bigcirc_X,\bigcirc_Y)$, $\neg\mathbf{P}(\bigcirc_X, \bigcirc_Y)$, $\mathbf{D}(\bigcirc_X, \bigcirc_Y)$, $\neg\mathbf{D}(\bigcirc_X, \bigcirc_Y)$, and $\mathbf{P}(\bigcirc_X,\overline{\bigcirc_Y})$. 

\begin{figure} 
  \centering  
\includegraphics[width=1\textwidth]{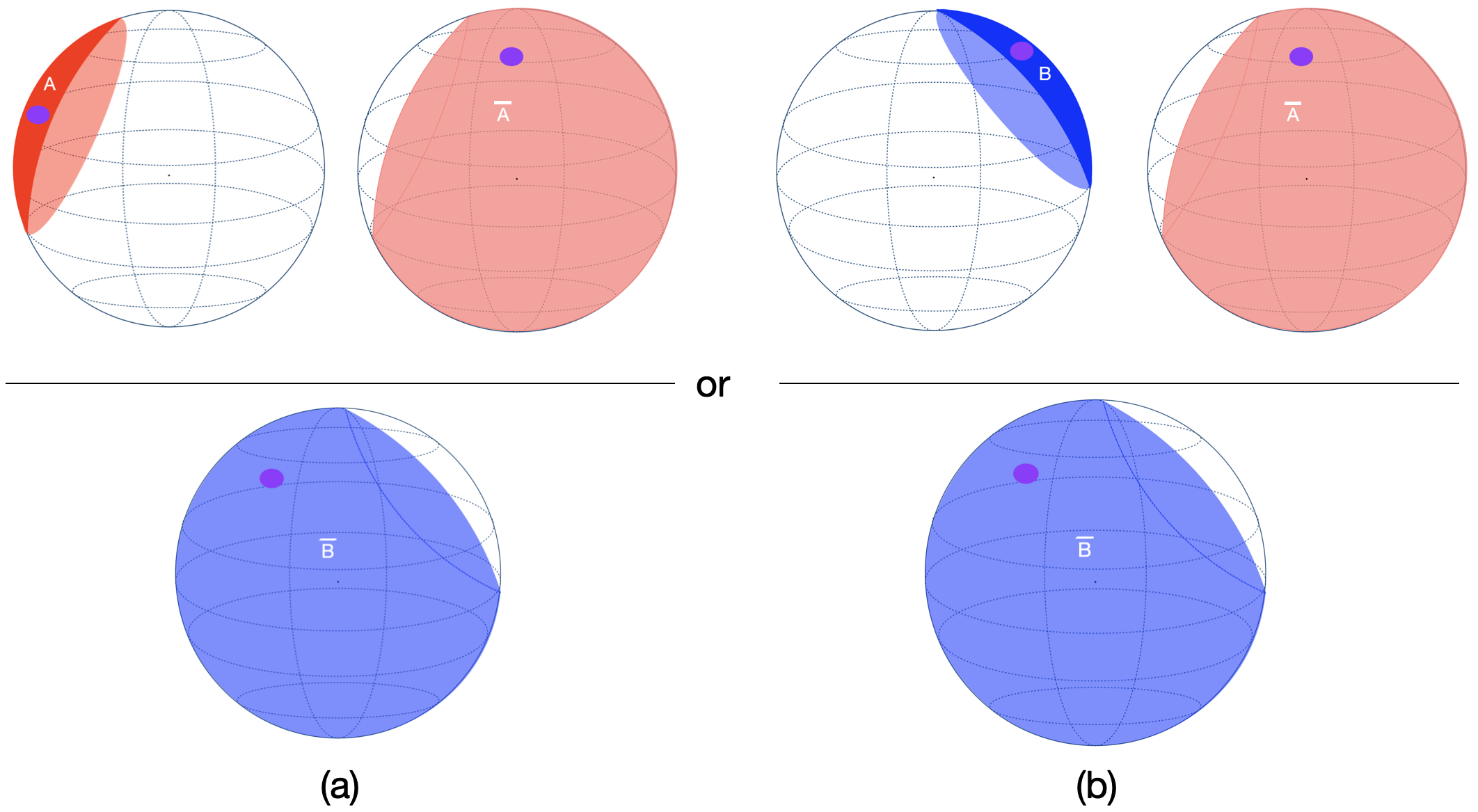}
  \caption{(a) Case 1: the grape is in jar A; (b) Case 2: the grape is in jar B. If our SphNN fails to construct a configuration for both cases, it will conclude that the grape in jar B is for sure.
  } 
  \label{monkey2}
\end{figure}

The main process of determining the satisfiability of syllogistic statements is a gradual descent process that begins with all circles coinciding. The configuration is updated iteratively by following a neighbourhood transition map. In the setting of constructing circle configurations on the surface of a sphere, the neighbourhood transition map share the same structure as the ones in ~\citep{djl2024sphere,djl2025}, with the condition that the sum of the diameters of two circles is less than the perimeter of the big circle of the sphere, as illustrated in Figure~\ref{monkey1}(a). This guarantees that our SphNN inherits the feature of \rfs of the original SphNN as follows. 
\begin{quote}
{\em For any satisfiable syllogistic statements, SphNN can correctly construct a sphere configuration as an Euler diagram at the global loss of zero in one epoch}.
\end{quote}
 So, if the SphNN fails to construct the target diagram after the first epoch, it will conclude that the input syllogistic statements are unsatisfiable, and the negation of the conclusion is valid (we outline the algorithm~\ref{algo} in the supplementary material). To determine whether jar B contains the grape, our SphNN shall refute the assumption that jar B does not contain the grape. This is achieved by failing to construct a circle configuration for this assumption, namely, $\mathbf{P}(\bigcirc_g,\bigcirc_A)\lor\mathbf{P}(\bigcirc_g,\bigcirc_B)$, $\mathbf{P}(\bigcirc_g, \overline{\bigcirc_A})$, and $\mathbf{P}(\bigcirc_g, \overline{\bigcirc_B})$. This is equivalent to two cases: (a) $\mathbf{P}(\bigcirc_g,\bigcirc_A)$, $\mathbf{P}(\bigcirc_g, \overline{\bigcirc_A})$, $\mathbf{P}(\bigcirc_g, \overline{\bigcirc_B})$, and (b) $\mathbf{P}(\bigcirc_g,\bigcirc_B)$, $\mathbf{P}(\bigcirc_g, \overline{\bigcirc_A})$, $\mathbf{P}(\bigcirc_g, \overline{\bigcirc_B})$, as illustrated in Figure~\ref{monkey2}. Our SphNN tries to construct a circle configuration for each case. After failing in both cases, it will confidently determine that the grape is in jar B.



\section{Experiments}
\label{exp}
We conducted four experiments to compare our SphNN and Euler Net with various training and testing datasets. In Experiment 1, we demonstrate that our SphNN achieves symbolic-level rigour in 16 syllogistic-style reasoning types~\citep{critical_thinking21}. In Experiment 2, we show that our SphNN keeps its rigour in classic syllogistic reasoning. In Experiment 3, we retrain Euler Net~\citep{WangJL18,WangJL20} for disjunctive syllogistic reasoning, reaching 100\% accuracy; In Experiment 4, we show that an Euler Net achieving 100.00\% in classic syllogistic reasoning can be trained to reach 100\% accuracy in disjunctive syllogistic reasoning. 
However, after that, its performance will drop in classic syllogistic reasoning, and subsequently drop when patterns of input images are different from those in the training data.

\subsection{Experiment 1}
\label{exp1}


\paragraph{Dataset} We translate 16 syllogistic-styled reasoning in Section~\ref{related_work} into a circle configuration in the disjunctive normal form (details are described in the supplementary material), and generate three other syllogistic conclusions, totalling 64 different circle configurations. Among 16 $\times$ 4 = 64 types of syllogistic reasoning statements, 32 types are valid. 

\paragraph{Method} 
To determine whether a syllogistic reasoning is valid, SphNN tries to construct a counter-example on the surface of a sphere. If successful, SphNN concludes that this syllogistic reasoning is invalid; otherwise, SphNN concludes that the original conclusion is valid.

\paragraph{Setup} 

The initial radius of a circle is $e^{-1}$. 
Three circles with radius $e^{-1}$ are randomly initialised as coinciding on the surface of a sphere with radius 1. We set the learning rate to 0.0001 and the maximum number of epochs $M=1$. All experiments were conducted on a MacBook Pro Apple M1 Max (10C CPU/24C GPU), 32 GB memory. We challenged SphNN to construct Poincaré spheres with the following dimensions $2, 3, 15, 30, 100, 200, 1000, 2000, 3000, 10000$.

\paragraph{Results} SphNN successfully determined 32 valid and 32 invalid syllogistic reasoning types by constructing circle configurations with dimensions from 2 to 10000, totalling 640 reasoning tasks. 
The mean time cost to determine a valid reasoning is $108.04$ seconds. The mean time cost to construct a counter-example for invalid reasoning is $7.07$ seconds; 221 among 320 ($69.06\%$) invalid reasoning cases are determined in less than $5$ seconds; 226 among 320 ($70.06\%$) valid cases are determined in less than $120$ seconds.

\subsection{Experiment 2}
\label{exp2}

\paragraph{Dataset} We use the dataset of SphNN~\citep{djl2025}.

\paragraph{Method and Setup} 
The same as in Experiment 1.

\paragraph{Results} Our SphNN successfully determined 24 valid and 232 invalid syllogistic reasoning types by constructing circle configurations with dimensions from 2 to 10000, totalling 2560 reasoning tasks. 
The mean time cost to determine a valid reasoning is $64.00$ seconds. The mean time cost to construct a counter-example for invalid reasoning is $11.27$ seconds; 2137 among 2320 ($92.11\%$) invalid reasoning cases are determined in less than $5$ seconds; 199 among 240 ($82.92\%$) valid cases are determined in less than $120$ seconds.

\subsection{A comparative study with supervised deep learning}
\label{en3}

\begin{figure}  
  \centering  
\includegraphics[width=1\textwidth]{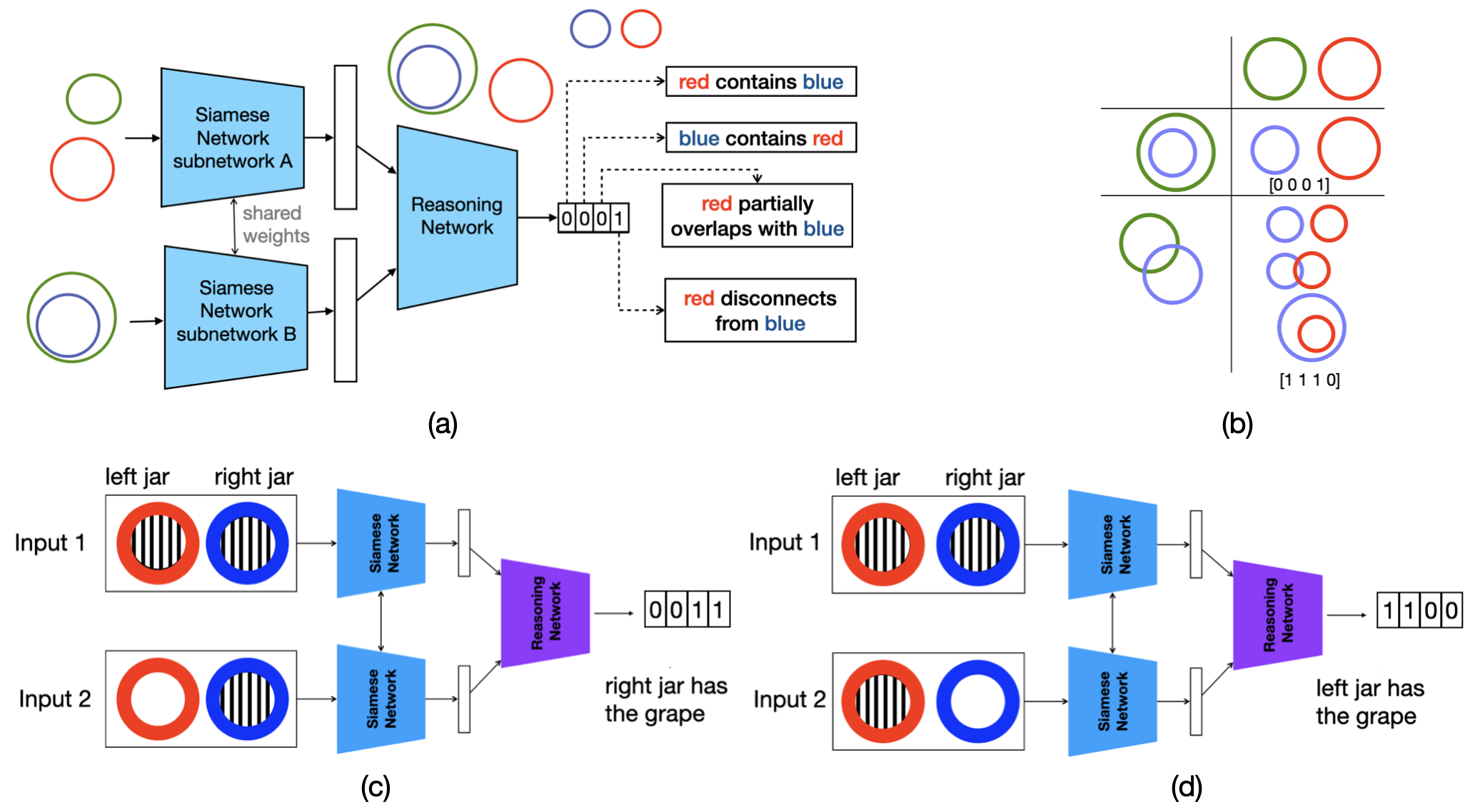}
  \caption{(a) The architecture of Euler Net; (b) The composition table of Euler Net. If the two premises are ``blue is inside green. green disconnects from red'', the combination result will only be ``blue disconnects from red'', represented as $[0, 0, 0, 1]$. (c, d)  Training data of Euler Net for disjunctive syllogistic reasoning.}
  \label{en}
\end{figure}

Developing neural syllogistic reasoning was extremely challenging and once considered utopian~\citep{laird2012}. 
Only recently were supervised neural networks, Euler Net, developed to approximate a substantial part of syllogistic reasoning \citep{WangJL18,WangJL20}, as illustrated in Figure~\ref{en}(a). 
The inputs of EN are two images, each consisting of two coloured circles with a set-theoretic relation. Colours of circles distinguish three terms in syllogistic reasoning. The common colour in the two input images is the middle term. With two Siamese networks, Euler Net encodes each input image into a latent vector. The output of EN is a vector representing the set-theoretic relation(s) between the subject and the predicate. The mapping from two premises to conclusions is enumerated in the combination table, where possible conclusions are symbolised as a vector, as illustrated in Figure~\ref{en}(b). The training data takes the form of ((image, image), vector). Euler Net can be trained to handle variants of classical syllogistic reasoning~\citep{WangJL18}. 

\subsubsection{Experiment 3}
\label{exp3}

\paragraph{Dataset} 
We repurpose Euler Net~\citep{WangJL18} for disjunctive syllogistic reasoning. The training data consists of two premises and a conclusion vector, representing the ``grape-in-jar" scenario. The first input image included two circles covered with stripes to represent uncertainty. The colour, thickness, and relative positioning of the circle were fixed, while the location of the centre was randomly varied. The second input image was identical to the first except that one circle was randomly chosen to be empty, indicating that it does not contain the grape. The output vector encoded the conclusion of the disjunctive reasoning: [1100] if the grape was in the left circle and [0011] if the grape was in the right circle, as illustrated in Figure~\ref{en}(c,d). 
\paragraph{Setup} 
Following the experiment settings in~\citep{WangJL18}, we created 96000 input-output records, among which 88000 records were used for training, 8000 records were used for validation and testing. 
\paragraph{Results} 
Euler Net achieved $100\%$ accuracy in disjunctive syllogistic reasoning on our testing dataset.  

\subsubsection{Experiment 4}
\label{exp4}
We examine whether the reasoning performance of Euler Net is affected by changes in input patterns and whether Euler Net can still achieve high performance in classic syllogistic reasoning.

\begin{figure} 
\includegraphics[width=1\textwidth]{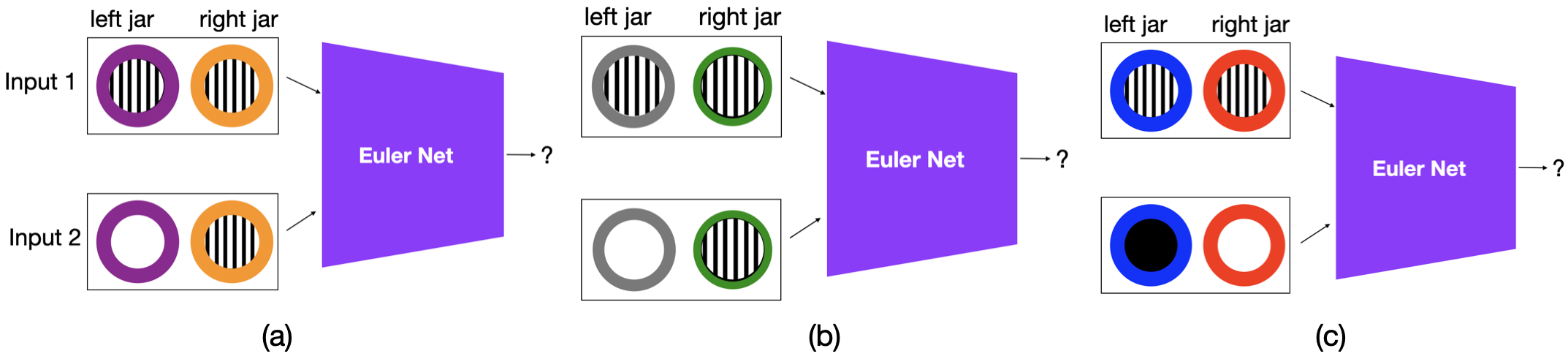}
\caption{Three variant testing datasets challenge the well-trained Euler Net in Experiment 3.}
\label{en4fl}
\end{figure}

\paragraph{Dataset} 
We designed three variant input patterns for disjunctive syllogistic reasoning, as shown in Figure~\ref{en4fl}(a-c). In the first variation, the colour of the circle was randomised;  In the second variation, both colour and thickness were randomised; In the third variation,  
the first input remained the same, but the second input depicted one circle as clear and the other as full. We generated testing data for classic syllogistic reasoning. The testing dataset for all experiments was the same as Experiment 3. 


\paragraph{Results} 
Accuracy decreased as the visual features varied: $75.00\%$ with randomised colour, $53.57\%$ with both colour and thickness variation, and $46.43\%$ with the alternative clear/full input pattern. For classic syllogistic reasoning, its performance dropped to 6.25\%.

\section{Conclusions and Outlooks}
\label{con}

Disjunctive syllogistic reasoning is a fundamental decision-making strategy that enables humans, animals, and AI systems to reach conclusions by eliminating incorrect candidates. In this paper, we compare three kinds of neural networks for basic decision-making, demonstrating the advantages of the method of {\em Reasoning through Explicit Model Construction}.
\begin{enumerate}
    \item This method constructs human-interpretable models. In contrast, supervised deep learning systems and LLMs are black boxes. 
    \item This method does not require training data, because premises contain all the information necessary to draw conclusions. In practice, this saves significant resources, including data, training time, and energy. Theoretically, this method completely eliminates the out-of-distribution problem.
    \item This method supports continuous learning -- a newly learned construction method can live with old ones. In contrast, supervised deep learning systems suffer severely from catastrophic forgetting. After Euler Net has learned disjunctive syllogistic reasoning, it will not be able to perform the already-learned classic syllogistic reasoning.
    \item This method constructs models by utilising a neighbourhood transition map -- a map designed for one task can be used for other tasks, as long as this map has the representation power to describe the target configuration. We show that the map designed for disjunctive syllogistic reasoning can be directly used for the other 15 syllogistic-style reasoning without any fine-tuning. In contrast, supervised deep learning systems need particular training data for each task. It may be possible for them to perform transfer learning with fine-tuning.  
    \item This method achieves symbolic-level syllogistic-styled logical reasoning. This paves a new path to resolve the conflict between symbolism and connectionism in the literature of Artificial Intelligence.
   
\end{enumerate}

Spheres are the intuitive visual tool for representing the part-whole relation, which is ubiquitous in everyday life and serves as a fundamental structural principle in taxonomic systems across the sciences~\citep{Smith96}. Many relational concepts can be reduced to the connection relation between regions, a tradition in philosophy and psychology dating back to~\citep{Laguna22a,whitehead29,Piaget54,Smith94,DongJPL}. For example, ``A is taller than B” can be defined as follows: Let X be a region above ground level. 
\begin{enumerate}
    \item A connects to X while standing on the ground; 
    \item B can not connect to X while standing on the ground.
\end{enumerate}
 Orientation can be reduced to distance comparisons~\citep{DongHans08}. The basic meaning of ``left” comes from proximity to the heart. ``B is to the left of C” can be interpreted as follows: If you project your body toward C, B will be closer to the side of your body where the heart is located. We can introduce fuzzy sphere boundaries to allow SphNN to represent graded membership and vague concepts~\citep{Zadeh65}.

A sphere is a generalisation of a vector by introducing a non-zero radius. Inversely, traditional neural networks can be viewed as special Sphere Neural Networks where all sphere radii are set to zero. Future research directions include seeking seamless integration of SphNNs with supervised deep learning and LLMs, applying them to cognitive modelling, and developing interpretable and safe AI models for high-stakes domains.

\bibliography{XBib,XBib_NN_s}
\bibliographystyle{apalike}

\appendix

\section{Additional experiments with GPT-5 family}

In Section~\ref{sintro}, we surveyed the recent evaluations of LLMs' reasoning performances in syllogistic reasoning, including disjunctive syllogism. Although these studies converge on the conclusion that LLMs still cannot perform perfect syllogistic reasoning, we may optimistically hope that they will be able to in the near future. Therefore, we conducted additional experiments to find an LLM and a matching input form that can achieve 100\% accuracy for classic syllogistic reasoning.  

We designed four input patterns as follows, totalling 256 $\times$ 4 = 1024 classic syllogistic reasoning tasks.
\begin{enumerate}
    \item Syllogistic statements with meaningful words, e.g. {\em all Greeks are humans};
    \item Syllogistic statements whose concepts are a string of two words connected by `\_', e.g. {\em some Greeks\_Mountain are not Humans\_Tomato};
    \item Syllogistic statements with simple symbols, e.g. {\em some S are P};
    \item Syllogistic statements with long random symbols, e.g. {\em some X2asfi3f23 are aH21ü01qH}. 
\end{enumerate}

We let OpenAI GPT-5 and GPT-5-nano decide whether a syllogistic reasoning is satisfiable and construct an Euler diagram if possible. 

The result is that GPT-5 only achieved 100\% accuracy in determining the satisfiability when fed with long random symbols. However, roughly 10\% of those correct decisions, GPT-5 created incorrect Euler diagrams. From GPT-5-nano to GPT-5, there were no significant improvements in limiting the number of correct decisions with wrong explanations. The training process of a neural network typically stops when it reaches 100\% accuracy, which indicates the ultimate unreliability of GPT-5's decision-making.
Details are listed in Table~\ref{gpt5sphnn}.

\begin{table}[h]
\caption{Syllogistic reasoning performance of OpenAI GPT-5-nano and GPT-5.  `expl' for correct explanation, `wrg' for wrong explanation. The `\#correct decision-wrg' column means a correct decision with a wrong explanation; the `\#wrong decision-expl' column means a wrong decision with a correct explanation.}
  \label{gpt5sphnn}
  \centering
  \begin{tabular}{cclllrr}
    \hline
 & &\multicolumn{2}{c}{\#correct decision}&\multicolumn{2}{c}{\#wrong decision}&\\
 \cline{3-4}\cline{5-6} version& surface form      & expl &  wrg & expl & wrg&\#simple acc\\
    \hline 
 gpt-5-nano& words &  160 ( 62.5\%)  & {\color{gray}{90}}    &{\color{gray}{5 }}  & {\color{gray}{1}}  & 97.7\%\\ 
 & double words &  230 ( 89.4\%)  & {\color{gray}{22}}    &{\color{gray}{4}}  & {\color{gray}{0}}  & 98.4\%\\
& simple symbols &  226 ( 88.3\%)  & {\color{gray}{24}}   
 &{\color{gray}{6}}  & {\color{gray}{0}}   & 97.7\%\\
& long random symbols &  222 ( 86.7\%)  & {\color{gray}{25}}   
 &{\color{gray}{9}}  & {\color{gray}{0}}  & 96.5\% \\
    \hline
gpt-5& words &  239 ( 93.4\%)  & {\color{gray}{16}}   
  &{\color{gray}{1}}  & {\color{gray}{0}}  &99.6\%\\ 
 & double words &  234 ( 91.4\%)  & {\color{gray}{21}}   &{\color{gray}{1}}  & {\color{gray}{0}}  & 99.6\%\\
& simple symbols &  236 ( 92.2\%)  & {\color{gray}{15}}   
  & {\color{gray}{5}}  & {\color{gray}{0}}  &98.0\% \\
& long random symbols &  231 ( 90.2\%)  & {\color{gray}{25}}   
&{\color{gray}{0}}  & {\color{gray}{0}}  & {\bf 100.0\%} \\
    \hline
  \end{tabular}
\end{table}

\section{Circle configurations for 16 Reasoning Types}

We listed circle configurations on the surface of a $n$-dimensional sphere as Euler diagrams for 16 syllogistic-styled reasoning types in \citep{critical_thinking21} as follows. 

\paragraph{1. Generalised modus ponens}  \syllogism{$\forall x F(x) \rightarrow G(x)$.\\ $F(a)$.}{$G(a)$.} 
Its circle configuration reads as ``Circle F is part of Circle G. Atomic Circle $a$ is part of Circle F. Atomic Circle $a$ is part of Circle G.'', written as follows.
\begin{center}
all F are G, all $a$  are F, all $a$  are G    
\end{center}
The other three syllogistic conclusions are as follows.
\begin{center}
all F are G, all $a$  are F, no $a$  are G 

all F are G, all $a$  are F, some $a$  are G 

all F are G, all $a$  are F, some $a$ are not G 
\end{center}

\paragraph{2. Negation variant of Generalised modus ponens} 
\syllogism{$\forall x F(x) \rightarrow \neg G(x)$.\\ $F(a)$.}{$\neg G(a)$.} 
Its circle configuration reads as ``Circle F is part of the complement Circle of Circle G. Atomic Circle $a$ is part of Circle F. Atomic Circle $a$ is part of the complement Circle of Circle G.'', written as follows.
\begin{center}
all F are c\_G, all $a$ are F, all $a$ are c\_G    
\end{center}
We use ``c\_'' to represent ``the complement Circle of''. The other three syllogistic conclusions are as follows.
\begin{center}
all F are c\_G, all $a$ are F, no $a$ are c\_G 

all F are c\_G, all a are F, some a are c\_G 

all F are c\_G, all $a$ are F, some $a$ are not c\_G 
\end{center}
\paragraph{3. Generalised contraposition}
\syllogism{$\forall x F(x) \rightarrow \neg G(x)$.}{$\forall x G(x) \rightarrow \neg F(x)$.}
Its circle configuration reads as ``Circle F is part of the complement Circle of Circle G. Circle G is part of the complement Circle of Circle G.'', written as follows.
\begin{center}
all F are c\_G, all G are c\_F    
\end{center}
The other three syllogistic conclusions are as follows.
\begin{center}
all F are c\_G, no G are c\_F 

all F are c\_G, some G are c\_F 

all F are c\_G, some G are not c\_F 
\end{center}
\paragraph{4. Negation variant of Generalised contraposition}
\syllogism{$\forall x F(x) \rightarrow G(x)$.}{$\forall x \neg  G(x) \rightarrow \neg F(x)$.}
Its circle configuration reads as ``Circle F is part of Circle G. The complement Circle of Circle F is part of the complement Circle of Circle F.'', written as follows.
\begin{center}
all F are G, all c\_G are c\_F    
\end{center}
The other three syllogistic conclusions are as follows.
\begin{center}
all F are G, no c\_G are c\_F 

all F are G, some c\_G are c\_F 

all F are G, some c\_G are not c\_F 
\end{center}
\paragraph{5. Hypothetical syllogism 1}
\syllogism{$\forall x F(x) \rightarrow G(x)$.\\ $\forall x G(x) \rightarrow H(x)$.}{$\forall x F(x) \rightarrow H(x)$.}
Its circle configuration reads as ``Circle F is part of Circle G. Circle G is part of Circle H. Circle F is part of Circle H'', written as follows.
\begin{center}
all F are G, all G  are H, all F are H    
\end{center}
The other three syllogistic conclusions are as follows.
\begin{center}
all F are G, all G  are H, no F  are H 

all F are G, all G  are H, some F are H 

all F are G, all G  are H, some F are not H 
\end{center}
\paragraph{6. Negation variant of Hypothetical syllogism 1}
\syllogism{$\forall x F(x) \rightarrow \neg G(x)$.\\ $\forall x \neg G(x) \rightarrow H(x)$.}{$\forall x F(x) \rightarrow H(x)$.}
Its circle configuration reads as ``Circle F is part of Circle G. Circle G is part of Circle H. Circle F is part of Circle H'', written as follows.
\begin{center}
all F are c\_G, all c\_G  are H, all F are H    
\end{center}
The other three syllogistic conclusions are as follows.
\begin{center}
all F are c\_G, all c\_G  are H, no F are H

all F are c\_G, all c\_G  are H, some F are H 

all F are c\_G, all c\_G  are H, some F are not H
\end{center}
\paragraph{7. Hypothetical syllogism 2}
\syllogism{$\forall x F(x) \rightarrow G(x)$.\\ $\forall x \neg H(x) \rightarrow \neg G(x)$.}{$\forall x F(x) \rightarrow H(x)$.}
Its circle configuration reads as ``Circle F is part of Circle G. The complement Circle of Circle H is part of the complement Circle of Circle G. Circle F is part of Circle H'', written as follows.
\begin{center}
all F are G, all c\_H  are c\_G, all F are H    
\end{center}
The other three syllogistic conclusions are as follows.
\begin{center}
all F are G, all c\_H  are c\_G, no F are H

all F are G, all c\_H  are c\_G, some F are H 

all F are G, all c\_H  are c\_G, some F are not H
\end{center}
\paragraph{8. Negation variant of Hypothetical syllogism 2}
\syllogism{$\forall x F(x) \rightarrow \neg G(x)$.\\ $\forall x \neg H(x) \rightarrow  G(x)$.}{$\forall x F(x) \rightarrow H(x)$.}
Its circle configuration reads as ``Circle F is part of the complement Circle of Circle G. The complement Circle of Circle H is part Circle of Circle G. Circle F is part of Circle H'', written as follows.
\begin{center}
all F are c\_G, all c\_H  are G, all F are H    
\end{center}
The other three syllogistic conclusions are as follows.
\begin{center}
all F are c\_G, all c\_H  are G, no F are H

all F are c\_G, all c\_H  are G, some F are H 

all F are c\_G, all c\_H  are G, some F are not H
\end{center}
\paragraph{9. Hypothetical syllogism 3}
 \syllogism{$\forall x F(x) \rightarrow G(x)$.\\ $\exists x  H(x) \rightarrow \neg G(x)$.}{$\exists x H(x) \rightarrow \neg F(x)$.}
Its circle configuration reads as ``Circle F is part of Circle G. An atomic Circle $a$ is part of Circle H. The atomic Circle $a$ is part of the complement Circle of Circle G. The atomic Circle $a$ is part of the complement Circle of Circle F'', written as follows.
\begin{center}
all F are G, all $a$ are H, all $a$ are c\_G, all $a$ are c\_F  
\end{center}
The other three syllogistic conclusions are as follows.
\begin{center}
all F are G, all $a$ are H, all $a$ are c\_G, no $a$ are c\_F

all F are G, all $a$ are H, all $a$ are c\_G, some $a$ are c\_F 

all F are G, all $a$ are H, all $a$ are c\_G, some $a$ are not c\_F
\end{center}
\paragraph{10. Negation variant of Hypothetical syllogism 3}
\syllogism{$\forall x \neg F(x) \rightarrow G(x)$.\\ $\exists x  H(x) \rightarrow \neg G(x)$.}{$\exists x H(x) \rightarrow F(x)$.}
Its circle configuration reads as ``The complement Circle of Circle F is part of Circle G. An atomic Circle $a$ is part of Circle H. The atomic Circle $a$ is part of the complement Circle of Circle G. The atomic Circle $a$ is part of of Circle F'', written as follows.
\begin{center}
all c\_F are G, all $a$ are H, all $a$ are c\_G, all $a$ are F  
\end{center}
The other three syllogistic conclusions are as follows.
\begin{center}
all c\_F are G, all $a$ are H, all $a$ are c\_G, no $a$ are F

all c\_F are G, all $a$ are H, all $a$ are c\_G, some $a$ are F 

all c\_F are G, all $a$ are H, all $a$ are c\_G, some $a$ are not F
\end{center}
\paragraph{11. Generalised modus tollens}
\syllogism{$\forall x F(x) \rightarrow G(x)$.\\ $\neg G(a)$.}{$\neg F(a)$.}
Its circle configuration reads as ``Circle F is part of Circle G. Atomic Circle $a$ is part of the complement Circle of  Circle F. Atomic Circle $a$ is part of the complement Circle of  Circle F.'', written as follows.
\begin{center}
all F are G, all $a$  are c\_G, all $a$  are c\_F   
\end{center}
The other three syllogistic conclusions are as follows.
\begin{center}
all F are G, all $a$  are c\_G, no $a$ are c\_F  

all F are G, all $a$  are c\_G, some $a$ are c\_F  

all F are G, all $a$  are c\_G, some $a$ are not c\_F 
\end{center}
\paragraph{12. Negation variant of Generalised modus tollens}

\syllogism{$\forall x F(x) \rightarrow \neg G(x)$.\\ $G(a)$.}{$\neg F(a)$.} 
Its circle configuration reads as ``Circle F is part of the complement Circle of Circle G. Atomic Circle $a$ is part of Circle F. Atomic Circle $a$ is part of the complement Circle of  Circle F.'', written as follows.
\begin{center}
all F are c\_G, all $a$  are G, all $a$ are c\_F   
\end{center}
The other three syllogistic conclusions are as follows.
\begin{center}
all F are c\_G, all $a$  are G, no $a$ are c\_F  

all F are c\_G, all $a$  are G, some $a$ are c\_F  

all F are c\_G, all $a$  are G, some $a$ are not c\_F 
\end{center}
\paragraph{13. Disjunctive syllogism}
\syllogism{$\forall x F(x) \rightarrow G(x)\lor H(x)$.\\ $\forall x  F(x) \rightarrow \neg G(x)$.}{$\forall x F(x) \rightarrow H(x)$.}
Its circle configuration reads as ``Circle F is part of Circle G or Circle H. Circle F is part of the complement Circle of  Circle G. Circle F is part of Circle H.'', written as follows.
\begin{center}
all F are G\_or\_H, all F are c\_G, all F are H   
\end{center}
The other three syllogistic conclusions are as follows.
\begin{center}
all F are G\_or\_H, all F are c\_G, no F are H  

all F are G\_or\_H, all F are c\_G, some F are H   

all F are G\_or\_H, all F are c\_G, some F are not H  
\end{center}
\paragraph{14. Negation variant of Disjunctive syllogism}
\syllogism{$\forall x F(x) \rightarrow G(x)\lor H(x)$.\\ $\forall x  G(x) \rightarrow \neg F(x)$.}{$\forall x F(x) \rightarrow H(x)$.}
Its circle configuration reads as ``Circle F is part of Circle G or Circle H. Circle G is part of the complement Circle of  Circle F. Circle F is part of Circle H.'', written as follows.
\begin{center}
all F are G\_or\_H, all G are c\_F, all F are H   
\end{center}
The other three syllogistic conclusions are as follows.
\begin{center}
all F are G\_or\_H, all G are c\_F, no F are H  

all F are G\_or\_H, all G are c\_F, some F are H    

all F are G\_or\_H, all G are c\_F, some F are not H  
\end{center}
\paragraph{15. Generalised dilemma}
\syllogism{$\forall x F(x) \rightarrow G(x)\lor H(x)$.\\ 
$\forall x  G(x) \rightarrow J(x)$.\\
$\forall x  H(x) \rightarrow J(x)$.}{$\forall x F(x) \rightarrow J(x)$.}
Its circle configuration reads as ``Circle F is part of Circle G or Circle H. Circle G is part of Circle J. Circle H is part of Circle J. Circle F is part of Circle J.'', written as follows.
\begin{center}
all F are G\_or\_H, all G are J, all H are J, all F are J   
\end{center}
The other three syllogistic conclusions are as follows.
\begin{center}
all F are G\_or\_H, all G are J, all H are J, no F are J  

all F are G\_or\_H, all G are J, all H are J, some F are J   

all F are G\_or\_H, all G are J, all H are J, some F are not J  
\end{center}
\paragraph{16. Negation variant of Generalised dilemma}
\syllogism{$\forall x F(x) \rightarrow G(x)\lor H(x)$.\\ 
$\forall x  J(x) \rightarrow \neg G(x)$.\\
$\forall x J(x) \rightarrow \neg H(x)$.}
{$\forall x F(x) \rightarrow \neg J(x)$.}
Its circle configuration reads as ``Circle F is part of Circle G or Circle H. Circle G is part of the complement Circle of Circle G. Circle J is part of the complement Circle of Circle J. Circle F is part of the complement Circle of Circle J.'', written as follows.
\begin{center}
all F are G\_or\_H, all J are c\_G, all J are c\_H, all F are c\_J   
\end{center}
The other three syllogistic conclusions are as follows.
\begin{center}
all F are G\_or\_H, all J are c\_G, all J are c\_H, no F are c\_J  

all F are G\_or\_H, all J are c\_G, all J are c\_H, some F are c\_J   

all F are G\_or\_H, all J are c\_G, all J are c\_H, some F are not c\_J 
\end{center}

\section{Algorithm}

Given a target circle configuration, we translate it into disjunctive normal form $f_1\lor\dots\lor f_m$. Each $f_i$ is a conjunctive form of circle relations. For each $f_i$, we search a looped chain of circles (in the chain, either a circle or its complement circle can appear, but not both), and let the SphNN determine whether this loop configuration of circles is satisfiable. If one $f_i$ is satisfiable, our SphNN will conclude $f_1\lor\dots\lor f_m$ satisfiable. For example, the configuration $\mathbf{P}(\bigcirc_A, \overline{\bigcirc_B})$, $\mathbf{P}(\overline{\bigcirc_B}, \bigcirc_C)$, and $\neg\mathbf{D}(\bigcirc_A, \bigcirc_C)$ is a relation among Circle A, the complement of Circle B, and Circle C. 
\begin{algorithm}
\DontPrintSemicolon 
\KwIn{A target circle configuration on an $n$-dimensional sphere.}
\SetKwData{loss}{Loss}
\SetKwData{circloop}{CircleLoop}
\KwOut{The input configuration is $\mathbf{Satisfiable}$ or $\mathbf{Unsatisfiable}$}  
Transform the input configuration into the disjunctive normal form $f_1\lor\dots\lor f_m$; \\
\loss$\gets +\infty$;\\ 

\ForEach{$f_i\in [f_1\dots f_m]$}
{
\circloop $\mbox{get a loop of circles from } f_i$;\\
\If{\circloop$==\emptyset$}
{$\mathbf{continue}$}
\Else{\loss$\gets\mbox{construct a configuration for \circloop}$;\\
\If{\loss$==0$}{$\mathbf{break}$}
}
}
\If{$\loss==0$}
{\Return{$\mathbf{Satisfiable}$}}
\Else{{\Return{$\mathbf{Unsatisfiable}$}}}
\caption{The main procedure of our SphNN}
\label{algo} 
\end{algorithm} 

\section{Code and Data}
Code and data are available upon request.

\end{document}